\relax
\documentclass[letterpaper]{article} 
\usepackage{aaai20}  
\usepackage{times}  
\usepackage{helvet}  
\usepackage{courier}  
\usepackage{url}  
\usepackage{graphicx}  
\usepackage[utf8]{inputenc}
\usepackage{subfigure}
\usepackage{booktabs} 
\usepackage{array}
\usepackage{amsfonts,amssymb}
\usepackage{algorithm,algpseudocode}
\algdef{SE}[DOWHILE]{Do}{doWhile}{\algorithmicdo}[1]{\algorithmicwhile\ #1}%
\PassOptionsToPackage{hyphens}{url}
\usepackage[export]{adjustbox}
\usepackage{xcolor}
\usepackage{tikz}
\usepackage{amsmath}

\definecolor{darkred}{rgb}{0.7,0,0}
\begin{document}
\title{Revisiting Semantic Representation and Tree \\Search for Similar Question Retrieval}

\author{
Tong Guo,
Huilin Gao
}
\maketitle
\begin{abstract}
This paper studies the performances of BERT combined with tree structure in short sentence ranking task. In retrieval-based question answering system, we retrieve the most similar question of the query question by ranking all the questions in datasets. If we want to rank all the sentences by neural rankers, we need to score all the sentence pairs. However it consumes large amount of time. So we design a specific tree for searching and combine deep model to solve this problem. We fine-tune BERT on the training data to get semantic vector or sentence embeddings on the test data. We use all the sentence embeddings of test data to build our tree based on k-means and do beam search at predicting time when given a sentence as query. We do the experiments on the semantic textual similarity dataset, Quora Question Pairs, and process the dataset for sentence ranking. Experimental results show that our methods outperform the strong baseline. Our tree accelerate the predicting speed by 500\%-1000\% without losing too much ranking accuracy.
\end{abstract}
\section{1 Introduction}

In retrieval-based question answering system \cite{wang2017bilateral,liu2018finding,guo2019deep}, we retrieve the answer or similar question from a large question-answer pairs. We compute the semantic similar score between question-question pairs or compute the semantic related score of question-answer pairs and then rank them to find the best answer. In this paper we discuss the similar question retrieval. For the similar question retrieval problem, when given a new question in predicting, we get the most similar question in the large question-answer pairs by ranking, then we can return the corresponding answer. We consider this problem as a short sentence ranking problem based on sentence semantic matching, which is also a kind of information retrieval task. 

Neural information retrieval has developed in several ways to solve this problem. This task is considered to be solved in two step: A fast algorithm like TF-IDF or BM25 to retrieve about tens to hundreds candidate similar questions and then the second step leverage the neural rankers to re-rank the candidate questions by computing the question-question pairs similarity scores. So one weakness of this framework with two steps above is that if the first fast retrieval step fails to get the right similar questions, the second re-rank step is useless. So one way to solve this weakness is to score all the question-question pairs by the neural rankers, however it consumes large amount of time. A full ranking may take several hours. See Fig 1. for the pipeline illustration.

\begin{figure}
\centering
\includegraphics[width=0.5\textwidth]{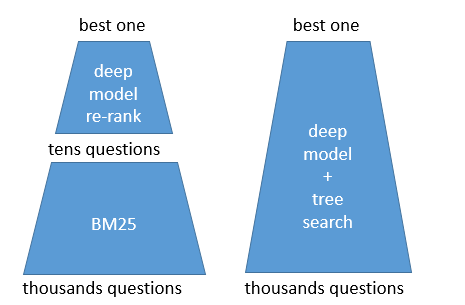}
\caption{The pipeline for retrieval-based question answering. The left is the classical pipeline and the right is our approach} \label{fig1}
\end{figure}

In this paper, to get the absolute most similar question on all the questions and solve the problem of long time for ranking all the data, inspired by the idea of \cite{zhu2018learning} and \cite{zhu2019joint}, we propose two methods: One is to compute all the semantic vector for all the sentence by the neural ranker offline. And then we encode the new question by the neural ranker online. Tree is an efficient structure for reducing the search space\cite{silver2016mastering}. To accelerate the speed without losing the ranking accuracy we build a tree by k-means for vector distance computation. Previous research \cite{qiao2019understanding,xu2019passage} shows that origin BERT\cite{devlin2018bert} can not output good sentence embeddings, so we design the cosine-based loss and the fine-tune architecture of BERT to get better sentence embeddings. Another method is to compute the similarity score by deep model during tree searching. In this paper, the words, distributed representations and sentence embeddings and semantic vector, are all means the final output of the representation-based deep model.

In summary our paper has three contributions: First, We fine-tuning BERT and get better sentence embeddings, as the origin embeddings from BERT is bad. Second, To accelerate the predicting speed, we build a specific tree to search on all the embeddings of test data and outperform the baseline. Third, after we build the tree by k-means, we search on the tree while computing the similarity score by interaction-based model and get reasonable results.

\section{2 Related Work}

In recent years, neural information retrieval and neural question answering research has developed several effective ways to improve ranking accuracy. Interaction-based neural rankers match query and document pair using attention-based deep model; representation-based neural rankers output sentence representations and using cosine distance to score the sentence pairs. There are many effective representation-based model include DSSM\cite{huang2013learning}, CLSM \cite{shen2014latent} and LSTM-RNN \cite{palangi2016deep} and many effective interaction-based model include DRMM\cite{guo2016deep} Match-SRNN\cite{wan2016match} and BERT\cite{devlin2018bert}.

Sentence embeddings is an important topic in this research area. Skip-Thought\cite{kiros2015skip} input one sentence to predict its previous and next sentence. InferSent\cite{conneau2017supervised} outperforms Skip-Thought. \cite{arora2016simple} is the method that use unsupervised word vectors\cite{pennington2014glove} to construct the sentence vectors which is a strong baseline. Universal Sentence Encoder \cite{cer2018universal} present two models for producing sentence embeddings that demonstrate good transfer to a number of other of other NLP tasks.

BERT is a very deep transformer-based\cite{vaswani2017attention} model. It first pre-train on very large corpus using the mask language model loss and the next-sentence loss. And then we could fine-tune the model on a variety of specific tasks like text classification, text matching and natural language inference and set new state-of-the-art performance on
them. However BERT is a very large model, the inference time is too long to rank all the sentence. 

We follow the BERT convention of data input format for encoding the natural language question. For single sentence classification task, the question $Q = \{w_1,w_2,...,w_n\}$ is encoded as following:

$[CLS], w_1, w_2, ..., w_n, [SEP]$

For sentence pair classification task, BERT passes two sentences to the transformer network and the target value is predicted. The question 1 $Q_1 = \{w_1,w_2,...,w_n\}$ and question 2 $Q_2 = \{w_1,w_2,...,w_m\}$ are encoded as following:

$[CLS], w_1, ..., w_n, [SEP], w_1, ..., w_m, [SEP] $

where [CLS] is a special symbol added in front of every input example, [SEP] is a special separator token, $n$, $m$ is the token number. Our fine-tune training follows the single sentence classification task convention for representation-based methods and follows the sentence pair classification task convention for interaction-based methods.
\section{3 Problem Statement and Approach}

\subsection{3.1 Problem Statement }
In this section, we illustrate the short sentence ranking task. In training time, we have a set of question pairs label by 1 for similar and by 0 for not similar. Our goal is to learn a classifier which is able to precisely predict whether the question pair is similar. But we can not follow the same way as sentence pair classification task of BERT, if we want to output the sentence embeddings for each of the sentence. In predicting time, we have a set of questions $Q = \{{q_1,q_2,...,q_n}\}$ that each have a labeled most similar question in the same set $Q$. Our goal is to use a question from the question set $Q$ as query and find the top N similar questions from the question set $Q$. Although the most similar question for the query is the one that we consider to be the most important one in question answering system, but the top N results may be applied to the scenario such as similar question recommendation. In the next section we describe our deep model and the tree building methods to solve this problem.

\begin{figure}
\centering
\includegraphics[width=0.5\textwidth]{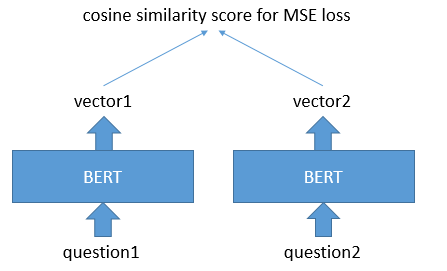}
\caption{The fine-tune training architecture of BERT for representation-based method} \label{fig2}
\end{figure}

\begin{figure}
\centering
\includegraphics[width=0.3\textwidth]{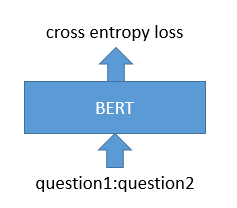}
\caption{The fine-tune training architecture of BERT for interaction-based method} \label{fig3}
\end{figure}

\subsection{3.2 Fine-tune Training}
In this subsection we describe our fine-tune methods for BERT. We call it representation-based method which fine-tune BERT to get sentence embeddings. We call it interaction-based method which fine-tune BERT to compute similarity score of sentence pairs during tree searching.

\subsubsection{Representation-based method}
 The sketch view is shown in Fig. 2. We input the two questions to the same BERT without concatenate them and output two vector representation. We adds a pooling operation to the output of BERT to derive a fixed sized sentence embedding. In detail, we use three ways to get the fixed sized representation from BERT: 

1. The output of the [CLS] token. We use the output vector of the [CLS] token of BERT for the two input questions. 

2. The mean pooling strategy. We compute mean of all output vectors of the BERT last layer and use it as the representation.

3. The max pooling strategy. We take the max value of the output vectors of the BERT last layer and use it as the representation.

Then the two output vectors from BERT compute the cosine distance as the input for mean square error loss:

$ loss = MSE(u \cdot v / (||u||*||v||),y) $

where $u$ and $v$ is the two vectors and $y$ is the label. The full algorithm is shown in Algorithm 1.

\subsubsection{Interaction-based method}
The fine-tune procedure is the same to the sentence pair classification task of BERT. The sketch view is shown in Fig. 3. Note that the colon in the figure denotes the concatenation operation. We concatenate the two questions to input it to BERT and use cross entropy loss to train. The full algorithm is shown in Algorithm 2. The fine-tuned model inputs the sentence in the tree node and query sentence as sentence pair to output the score.

\begin{algorithm}
\caption{Pipeline for representation-based BERT}
\label{alg:A}
\begin{algorithmic}[1]
\State{init BERT model BERT-A}
\For{epoch $\in$ epoch\_num}
\For{question\_pairs $\in$ train\_question\_pairs}
\State{fine-tune BERT-A to BERT-B}
\EndFor
\EndFor
\State{all\_embeddings = set()}
\For{question $\in$ test\_questions}
\State{question\_embedding=BERT-B.forward(question)}
\State{all\_embeddings.add(question\_embedding)}
\EndFor
\State{use all\_embeddings to init the tree T}

\For{question $\in$ test\_questions}
\State{question\_embedding=BERT-B.forward(question)}
\State{result=T.beam\_search(question\_embedding, topN)}
\State{eval(result,true\_rank)}
\EndFor

\end{algorithmic}
\end{algorithm}

\begin{algorithm}
\caption{Pipeline for interaction-based BERT}
\label{alg:A}
\begin{algorithmic}[1]
\State{init BERT model BERT-A}
\For{epoch $\in$ epoch\_num}
\For{question\_pairs $\in$ train\_question\_pairs}
\State{fine-tune BERT-A to BERT-B on Fig 2.} 
\State{fine-tune BERT-A to BERT-C on Fig 3.}
\EndFor
\EndFor
\State{all\_embeddings = set()}
\For{question $\in$ test\_questions}
\State{question\_embedding=BERT-B.forward(question)}
\State{all\_embeddings.add(question\_embedding)}
\EndFor
\State{use test\_questions and all\_embeddings to init the tree T}
\State{use nearest questions of cluster centers to init the T's non-leaf node}
\For{question $\in$ test\_questions}
\State{result=T.beam\_search(question, BERT-C, topN)}
\State{eval(result,true\_rank)}
\EndFor

\end{algorithmic}
\end{algorithm}

\begin{figure}
\centering
\includegraphics[width=0.5\textwidth]{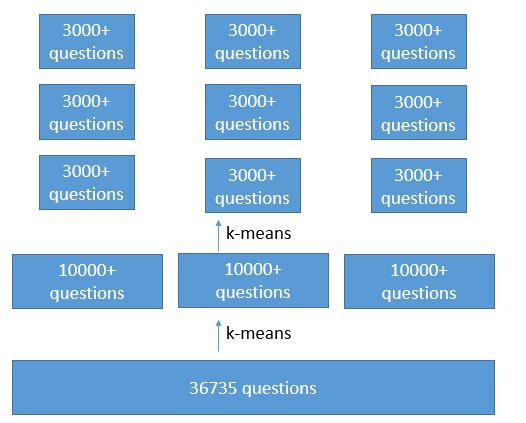}
\caption{The k-means clustering for building the tree with cluster number = 3} \label{fig3}
\end{figure}

\begin{figure}
\centering
\includegraphics[width=0.5\textwidth]{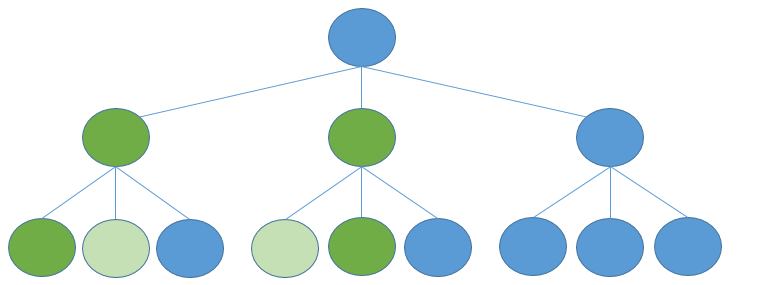}
\caption{The beam search strategy (beam size = 2): deep green nodes are the final choices and light green nodes are the candidate nodes} \label{fig4}
\end{figure}

\subsection{3.3 Tree Building}

In this section we describe our tree building strategies. In our tree, each non-leaf node have several child nodes. The leaf nodes contain the real data and the non-leaf nodes are virtual but undertake the function for searching or undertake the function that lead the path to the leaf nodes. 

\begin{table*}
\caption{Our 5-K tree result compare to the baseline}\label{tab1}
\centering
\begin{tabular}{|l|l|l|l|l|l|}

\hline
Methods & MAP & P@1 & MRR & NDCG & MRR@10\\
\hline
wordvec\cite{arora2016simple} & 0.274 & 0.232 & 0.275 & 0.305 & 0.272 \\
\hline
InferSent & 0.442 & 0.373 & 0.443 & 0.479 & 0.442 \\
\hline
Universal Sentence Encoder & 0.485 & 0.425 & 0.487 & 0.522 & 0.485 \\
\hline
our interaction-based method & 0.454 & 0.384 & 0.455 & 0.483 & 0.452 \\ 
\hline
our representation-based method ([CLS]) & 0.557 & 0.478 & 0.560 & 0.607 & 0.557 \\ 
\hline
our representation-based method (mean) & 0.570 & 0.493 & 0.574 & 0.619 & 0.571 \\ 
\hline
our representation-based method (max) & 0.564 & 0.488 & 0.569 & 0.614 & 0.565 \\ 
\hline
\end{tabular}
\end{table*}

\begin{table*}
\caption{Analysis of ranking accuracy losing for representation-based method}\label{tab5}
\centering

\begin{tabular}{|l|l|l|l|l|l|}
\hline
Methods & MAP & P@1 & MRR & NDCG & MRR@10\\
\hline
our 10-K tree  & 0.551 & 0.474 & 0.557 & 0.598 & 0.555 \\ 
\hline
our 8-K tree & 0.555 & 0.478 & 0.560 & 0.603 & 0.560 \\ 
\hline
our 5-K tree & 0.570 & 0.493 & 0.574 & 0.619 & 0.571 \\ 
\hline
k-d tree   & 0.610 & 0.523 & 0.614 & 0.668 & 0.611 \\ 
\hline
compute-all (cosine)  & 0.610 & 0.523 & 0.614 & 0.668 & 0.611 \\ 
\hline
compute-all (euclidean) & 0.610 & 0.523 & 0.614 & 0.668 & 0.611 \\ 
\hline

\end{tabular}
\end{table*}

\subsubsection{Representation-based method}

After all the embeddings of test data are computed, we start to build the tree by k-means. The sketch figure for tree building is shown in Fig. 4. In real the child nodes for each parent may be not that balance. We cluster the embeddings recursively. The sentence embeddings are all in the leaf nodes. The non-leaf node representation is important for the tree search as they pave the way and lead to the right leaf nodes. We use the k-means clustering centers as the non-leaf node embeddings. We think the clustering centers is a good solution for the non-leaf node representation, as it is hard to get the exact representation from the child nodes for the parent nodes. As we already get all the embeddings of test data, we only need to compute the vector distance during tree searching.

\subsubsection{Interaction-based method}
For interaction-based BERT, we first build the tree by sentence embeddings from the representation-based method above and then use the sentence strings as the leaf nodes. We take the nearest 1-5 sentence strings of cluster centers for the non-leaf node. This strategy has been proved to be effective in experiments.

\subsection{3.4 Tree Search}

In this section we describe our tree searching strategies. The two strategies are almost the same. The difference is that representation-based method compute the vector distance at each node but interaction-based method use the deep model to score the string pair at each node.

\subsubsection{Representation-based method}
At predicting time, we use beam search from top to down to get the nearest top N vectors for the given query vector from the whole tree. If we set the beam size to N, we first choose the top N nodes from the all the child nodes of first level and then search among the chosen child nodes' child nodes for the second level. Then we choose top N nodes from the second level. The detail beam search strategy is shown in Fig 5. 

\subsubsection{Interaction-based method}
At predicting time, we compute the score of two sentences by BERT for each node while we are searching the tree. As we take 1-5 sentence for a non-leaf node, we use the max similarity score to decide which non-leaf node is better. The detail beam search strategy is the same as Fig 5. shows. The more sentences that are nearest to the clustering centers we take for one non-leaf node, the more computation time we need to do for a non-leaf node. But the most computation time is consumed at the leaf nodes as leaf node number is much larger than non-leaf node number.

\section{4 Experiments}
In this section, we describe the datasets, experiments parameter detail and the experimental result. Then, we give a detailed analysis of the model and experiment results.

\begin{table*}
\caption{Case study for query: Who is the best bodybuilder of all time ?}\label{tab5}
\centering

\begin{tabular}{|l|l|l|}
\hline
Methods & result & Label \\
\hline
our tree, rank 1 & How much money do professional strongmen make ? & 0  \\ 
\hline
our tree, rank 2 & Why did Indians want Independence from Britain ? & 0   \\ 
\hline
our tree, rank 3  & Do you think the Indian marriage traditions needs a change ? & 0  \\ 
\hline
our tree, rank 4  & Will India still able to win gold medal at Rio Olympics even after 4 days and no medal ? & 0  \\ 
\hline
our tree, rank 5  & I 'm 18 and have started to do weight lifting . Will it stop my height to increase ? & 0  \\ 
\hline
compute-all, rank 1  & Who is the best bodybuilder ? & 1   \\
\hline
compute-all, rank 2 & Who is the most skillful fighter in Game of Thrones ? & 0  \\ 
\hline
compute-all, rank 3 & Which is the best website maker for an online shop ? & 0  \\ 
\hline
compute-all, rank 4 & Where do you buy kratom online ? & 0  \\ 
\hline
compute-all, rank 5 & Who are Grubwithus competitors ? & 0  \\ 
\hline

\end{tabular}
\end{table*}

\subsection{4.1 Data Description}
We evaluate the performance on the Quora Question Pairs datasets. Based on the Quora Question Pairs datasets, we combine the dev data and test data to get a dataset of 20000 question pairs, which contains 10000 pairs with label 1 and 10000 pairs with label 0.  After remove the duplicate questions, we get a datasets of 36735 questions. We compute the all embeddings for the 36736 questions in advance. And then we use the 10000 questions which have label 1 as 10000 queries. For each query it compute 36735 cosine distances if we loop all the 36735 questions. We take the top 20 questions for the evaluation of ranking. The training datasets is 384348 question pairs.

\subsection{4.2 Fine-tune Training}
We use the pre-trained BERT-base model file from here\footnote[1]{\url{https://github.com/google-research/bert}}. The max sequence length is 64 and the batch size is 32. The hidden dimension of BERT or output representation dimension is 768. We use Adam optimizer with learning rate 2e-5, and a linear learning rate warm-up over 10\% of the training data. 

\subsection{4.3 Tree Building}
We choose 5,8,10 as clustering number for k-means. We name the trees 5-K tree, 8-K tree and 10-K tree, based on the clustering number. The depth for the tree is 5 levels for 36735 vectors. In predicting time, the 5-K tree is the slowest with best accuracy tree and the 10-K tree is the fastest with worst accuracy tree. The 8-K tree is in the middle of them.

\subsection{4.4 Results}
We evaluate the retrieved top N sentences by Mean Average Precision (MAP), Precision @ 1 (P@1), Normalized Discounted Cumulative Gain (NDCG), Mean Reciprocal Rank (MRR) and MRR@10. The \cite{arora2016simple} baseline is from here\footnote[2]{\url{https://github.com/peter3125/sentence2vec}} and the \cite{conneau2017supervised} baseline if from here\footnote[3]{\url{https://github.com/facebookresearch/InferSent}}. The detail compare result is shown in Table 1. and Table 2. The compute-all result means we score all the vector pairs from 0 to end sequentially. The vector distance computation of compute-all uses cosine distance and euclidean distance, and k-d tree uses euclidean distance. The speed comparison is shown in Table 4. We count the number of vector distance computation times for representation-based method or the number of scoring times for sentence pair for interaction-based method. Our tree-based methods outperform \cite{arora2016simple} by 113\%, outperform \cite{conneau2017supervised} by 32\% and outperform \cite{cer2018universal} by 16\% in the top 1 accuracy.

\subsection{4.5 Case Study and Error Analysis}
We show some examples from the eval results to demonstrate the ability of our methods.
Table 3 shows the retrieval result of top 5 for the query question "Who is the best bodybuilder of all time ?" for compute-all and our 10-K tree. The results show that the ranking accuracy losing may be caused by the non-leaf representation's error, as the results of our tree is far from the query question. We even can not find the right result in the retrieved top 20 questions. We think the non-leaf node lead to the wrong children in tree searching. It is the weakness of our tree building strategy. 

\begin{table}
\caption{vector distance computation times to retrieve top 20 for 36735 pairs in predicting}\label{tab3}
\centering

\begin{tabular}{|l|l|}
\hline
Methods & times\\
\hline
our 5-K tree & 6000-7000 \\ 
\hline
our 8-K tree & 3000-4000 \\ 
\hline
our 10-K tree & 2000-3000 \\ 
\hline
k-d tree & about 24000  \\ 
\hline
compute-all & 36735  \\ 
\hline

\end{tabular}
\end{table}
\section{5 Conclusion and Future Work}
In this paper, we study the problem of short sentence ranking for question answering. In order to get best
similar score in all the questions when given a question as query and accelerate the predicting speed, we propose two methods. The first method is compute the representation for all the questions in advance and build a tree by k-means. The second method is to train a deep model and then use it to compute similarity scores of two sentences during tree searching. The experimental results show that our methods outperform the strong baseline on the short sentence retrieval datasets we construct. The sentence embeddings quality may be improved by better BERT\cite{liu2019roberta} or the XLNet\cite{yang2019xlnet} and we will discover more powerful non-leaf node embeddings for the tree search and evaluate on other datasets\cite{cer2017semeval}, as previous research \cite{zhu2018learning,zhu2019joint} shows that the tree's preformance could reach the performance of compute-all. In conclusion, our goal is to discover better embeddings and better tree structure in the future.

\bibliographystyle{aaai}
\bibliography{ref1}
\end{document}